\documentclass[conference]{IEEEtran}
\IEEEoverridecommandlockouts
\usepackage{cite}
\usepackage{amsmath,amssymb,amsfonts}
\usepackage{algorithmic}
\usepackage{graphicx}
\usepackage{subfigure} 
\usepackage{titlesec}
\usepackage{textcomp}
\usepackage{xcolor}
\def\BibTeX{{\rm B\kern-.05em{\sc i\kern-.025em b}\kern-.08em
    T\kern-.1667em\lower.7ex\hbox{E}\kern-.125emX}}
\usepackage{titlesec}
\usepackage{bm}

\begin{document}

\title{TSGN: Temporal Scene Graph Neural Networks 
with Projected Vectorized Representation for Multi-Agent Motion Prediction} 
{\footnotesize}
\author{
\IEEEauthorblockN{Yunong Wu$^{1}$, Thomas Gilles$^{2}$, Bogdan Stanciulescu$^{2}$, Fabien Moutarde$^{2}$}
\IEEEauthorblockA{$^{1}$ School of Computation, Information and Technology, Technical University of Munich}
\IEEEauthorblockA{$^{2}$ MINES ParisTech, PSL University, Center for robotics}
\IEEEauthorblockA{\{wuyun\}@in.tum.de , \{thomas.gilles, bogdan.stanciulescu, Fabien.Moutarde\}@minesparis.psl.eu}}

\maketitle
\begin{abstract}
Predicting future motions of nearby agents is essential for an autonomous vehicle to take safe and effective actions. In this paper, we propose TSGN, a framework using Temporal Scene Graph Neural Networks with projected vectorized representations for multi-agent trajectory prediction. Projected vectorized representation models the traffic scene as a graph which is constructed by a set of vectors. These vectors represent agents, road network, and their spatial relative relationships. All relative features under this representation are both translation- and rotation-invariant. Based on this representation, TSGN captures the spatial-temporal features across agents, road network, interactions among them, and temporal dependencies of temporal traffic scenes. TSGN can predict multimodal future trajectories for all agents simultaneously, plausibly, and accurately. Meanwhile, we propose a Hierarchical Lane Transformer for capturing interactions between agents and road network, which filters the surrounding road network and only keeps the most probable lane segments which could have an impact on the future behavior of the target agent. Without sacrificing the prediction performance, this greatly reduces the computational burden. Experiments show TSGN achieves state-of-the-art performance on the Argoverse motion forecasting benchmark.
\end{abstract}

\section{INTRODUCTION}\label{chapter:INTRODUCTION}
Predicting the motion of multiple agents is an essential step in autonomous driving. It helps autonomous driving vehicles to plan their actions and prevent  accidents. However, the traffic scene is highly complex. It contains agents, road network, and interactions among them. The prediction model needs to take these entities as inputs and outputs reasonable multimodal trajectories that the target agent could take in the future. 
The traffic scene is viewed differently from the perspective of different agents. However, most exsiting methods select one target agent each time and take it as the central reference to perform the coordinate transformation on the entire scene, which is not symmetric for the other agents and results in only one prediction at a time. It is inefficient and not robust to the coordinate transformation. To alleviate these problems, HiVT \cite{zhou2022hivt} takes a symmetric way to encode the traffic scene. It divides the traffic scene into a set of local regions, each corresponding to one agent and centered on it, and uses a local-global mechanism to encode agents' features.




We adopt the symmetrical representation of HiVT and its local-global mechanisms. Based on them, we propose an extended scene representation method: Projected vectorized representation and an optimized prediction framework: TSGN. Our representation method introduces more features about agents and lane segments. It projects all relative spatial features between entities into the direction of the target entity for every timestamp, such that, all these relative features under this representation are also both translation- and rotation-invariant. TSGN consists of diverse interaction modules and temporal dependencies modules. These modules are constructed based on the transformer structure. TSGN treats the traffic scene in its entirety and encodes the agents’ dynamics, surrounding lane segments, and interactions among them across time steps. In addition, we present a Hierarchical Lane Transformer module inside TSGN to capture the influence of the road network on the future motion of target agents. Unlike most approaches that use the full structural information of the surrounding road network, the Hierarchical Lane Transformer only selects the most probable lane segments which could have an impact on
the future behavior of the target agent. It greatly reduces the computation burden and enables the model to do faster predictions without sacrificing the prediction performance.

Our contributions are summarized as follow:
\begin{itemize}
\item We extend the scene representation method of HiVT by adding new features about agents, lane segments, and their spatial relative relationships. We adopt a projected vectorized representation for all relative features of the traffic scene for every timestamp, which describes the relative spatial relationships in a more detailed way.
\item We propose a Hierarchical Lane Transformer which only selects lane segments that affect the future behaviors of target agents the most for modeling the interactions, enabling much faster and equally accurate predictions. 
\item Our designed TSGN can make plausible and accurate predictions. We validate the performance of TSGN on Argoverse forecasting benchmark. 
It ranked $1^{st}$ in terms of minADE on the leaderboard on September 13, 2022.
\end{itemize}

\section{RELATED WORK}\label{chapter:PRELIMINARIES}



Recently deep learning-based models have brought great progress to the motion prediction. According to the different representation methods, these learning-based models can be divided into rasterized-representation-based models and vectorized-representation-based models.
Rasterized representation of traffic scenes has benefits of unitedly rendering various types of input information (e.g., HD map, agents' trajectories, spatial relationships) into an image with multiple channels. In \cite{djuric2020uncertainty, cui2019multimodal}, different types of entities in traffic scenes are assigned different colors. HOME \cite{gilles2021home} rasterizes elements of HD map in different semantic channels and adds historical trajectories of the target agent and other agents on multiple channels to represent agents' physical states at each timestamp. In the same branch, \cite{hong2019rules, chai2019multipath} adopt a similar rasterized representation method. However, rasterized representation methods are limited to restricted image size and CNNs have constrained receptive fields. Besides, they cannot capture the complex structural information of HD map and it is difficult for them to represent the agents' dynamics. 

More recently, vectorized representation methods are proposed to capture the structural information of traffic scenes. They use graph neural networks (GNNs) to capture the spatial features of agents and HD map and to learn the relationships between them. VectorNet \cite{gao2020vectornet} is the first to use the vectorized representation method to encode both HD map and agents' dynamics. TNT \cite{zhao2021tnt} and DenseTNT \cite{gu2021densetnt} adopt the same representation method and use VectorNet as their backbone network. LaneGCN \cite{liang2020learning} treats lane segments as nodes and constructs a lane graph from the lane segments. \cite{gilles2021thomas, gilles2022gohome, zeng2021lanercnn, deo2022multimodal} adopt similar lane graph representation methods as LaneGCN. TPCN \cite{ye2021tpcn} and its extension DCMS \cite{ye2022dcms} treat trajectory prediction problems as joint temporal point cloud learning and use PointNet++ \cite{qi2017pointnet++} in the point representation learning. HDGT \cite{jia2022hdgt} models the traffic scene as a heterogeneous graph with different types of nodes and edges. The approach most related to this paper is HiVT \cite{zhou2022hivt}, which uses a translation-invariant scene representation method that avoids using absolute positions and characterizes the geometric entities with relative positions and constructs rotation-invariant transformers to model the different interactions between the vectorized entities locally and globally.
\section{TEMPORAL TRAFFIC SCENE REPRESENTATION}\label{chapter:TEMPORAL TRAFFIC SCENE REPRESENTATION}
We adopt the feature representation of HiVT and extend it. In this section, we introduce our feature representation in details. Our introduced extended features of the traffic scene and the original features of HiVT in their definitions are presented in blue and black respectively. 


\subsection{Node Feature Representation}
For traffic agents,  we extract their trajectory segments which are in a form of directed splines with respect to time. We consider each agent at each timestamp as a node and we characterize these temporal geometric nodes' attributes as $\mathbf{n}_{a} = \{{\mathbf{R}^{T}_{i}}^{\top}\mathbf{d}^{t}_i, {\color{cyan}{\mathbf{R}^{T}_{i}}^{\top}\mathbf{v}^{t}_i}, {\color{cyan}s^{t}_i}, {\color{cyan}{\mathbf{R}^{T}_{i}}^{\top}\mathbf{a}^{t}_i}, {\color{cyan}\Delta t^{t}}, \mathbf{b}_i|i=1,...,N_{t}, t=0,...,19\}$ with
\begin{align}
    &\mathbf{p}^{t}_i = [x^{t}_i, y^{t}_i] & &{\color{cyan}s^{t}_i} = \lVert {\color{cyan}\mathbf{v}^{t}_i} \rVert\\
    &\mathbf{d}^{t}_i = \mathbf{p}^{t}_i - \mathbf{p}^{t-1}_i & &\alpha^{t}_i = \arctan{\frac{{d_y}^{t}_i}{{d_x}^{t}_i}}\\
    &{d_x}^{t}_i = x^{t}_i - x^{t-1}_i & &{\color{cyan}\mathbf{a}^{t}_i} = [\cos(\alpha^{t}_i),\sin(\alpha^{t}_i)]\\
    &{d_y}^{t}_i = y^{t}_i - y^{t-1}_i & &{\color{cyan}\Delta t^{t}} = t^{t} - t^{t-1}\\
    &{\color{cyan}\mathbf{v}^{t}_i} = \frac{\mathbf{d}^{t}_i}{\Delta t^{t}}
\end{align}
where $N_{t}$ represents the total number of agents at timestamp $t$,  $\mathbf{p}^{t}_i$ is the location vector which consists of the lateral and longitudinal position of agent $i$ at timestamp $t$, $\mathbf{d}^{t}_i$ is the displacement vector from timestamp $t-1$ to timestamp $t$, $\mathbf{v}^{t}_i$ is the velocity, $s^{t}_i$ is the speed, $\mathbf{a}^{t}_i$ is the heading vector which consists of the cosine and sine of the agent's heading $\alpha^{t}_i$, $\Delta t^{t}$ is the timestamp difference, $\mathbf{R}^{T}_{i}$ is the rotation matrix parameterized by the heading $\alpha^{\top}_{i}$ at the current timestamp $T$ ($19$), and $\mathbf{b}_i$ corresponds to semantic attribute. The sampling rate is $10$Hz, but we found it deviates in different scenes, thus we include the timestamp difference into the node feature.  

For road network, we extract coordinates of lane segments and their semantic attributes, such as turn directions. We similarly vectorize lane segments as the vectorization for agents. The geometric node attributes of lane segments $\mathbf{n}_{l} = \{\mathbf{d}_\xi,{\color{cyan}\mathbf{a}_\xi},{\color{cyan}l_\xi},\mathbf{b}_\xi|\xi=1,...,M\}$ consist of displacement vector $\mathbf{d}_\xi$ of each lane segment, heading vector $\mathbf{a}_\xi$, length of the lane segment $l_\xi$, and also the semantic attribute $\mathbf{b}_\xi$, where
\begin{align}
    &\mathbf{d}_\xi = \mathbf{p}^1_\xi  - \mathbf{p}^0_\xi &{\color{cyan}l_\xi} = \lVert \mathbf{d}_\xi \rVert\\
    &{\color{cyan}\mathbf{a}_\xi} = [\cos(\alpha_\xi),\sin(\alpha_\xi)]
\end{align}
$M$ is the total number of lane segments, $\mathbf{p}^0_\xi$ and $\mathbf{p}^1_\xi$ represent the start point and end point of the lane segment $\xi$ respectively. Instead of using absolute representations, using relative representations ensure these state quantities are not affected by coordinate translation transformations. 
\subsection{Edge Feature Representation} 
The node features do not preserve relative spatial relationships. Thus, we further introduce the geometric edge attributes between entities. We characterize the edge attributes between agents as $\mathbf{e}_{aa} = \{{\mathbf{R}^{T}_{i}}^{\top}\mathbf{d}^{t}_{ij}, {\color{cyan}{\mathbf{R}^{T}_{i}}^{\top}\mathbf{v}^{t}_{ij}}, {\color{cyan}l_{ij}}, {\color{cyan}s_{ij}},{\color{cyan}\mathbf{d}^{t}_{j2i}},{\color{cyan}\mathbf{v}^{t}_{j2i}},{\color{cyan}\mathbf{a}^{t}_{j2i}}|t=0,...,19;i,j=1,...,N_t, i\neq j\}$, where
\begin{figure}[htbp] 
\centering
\includegraphics[height=4cm]{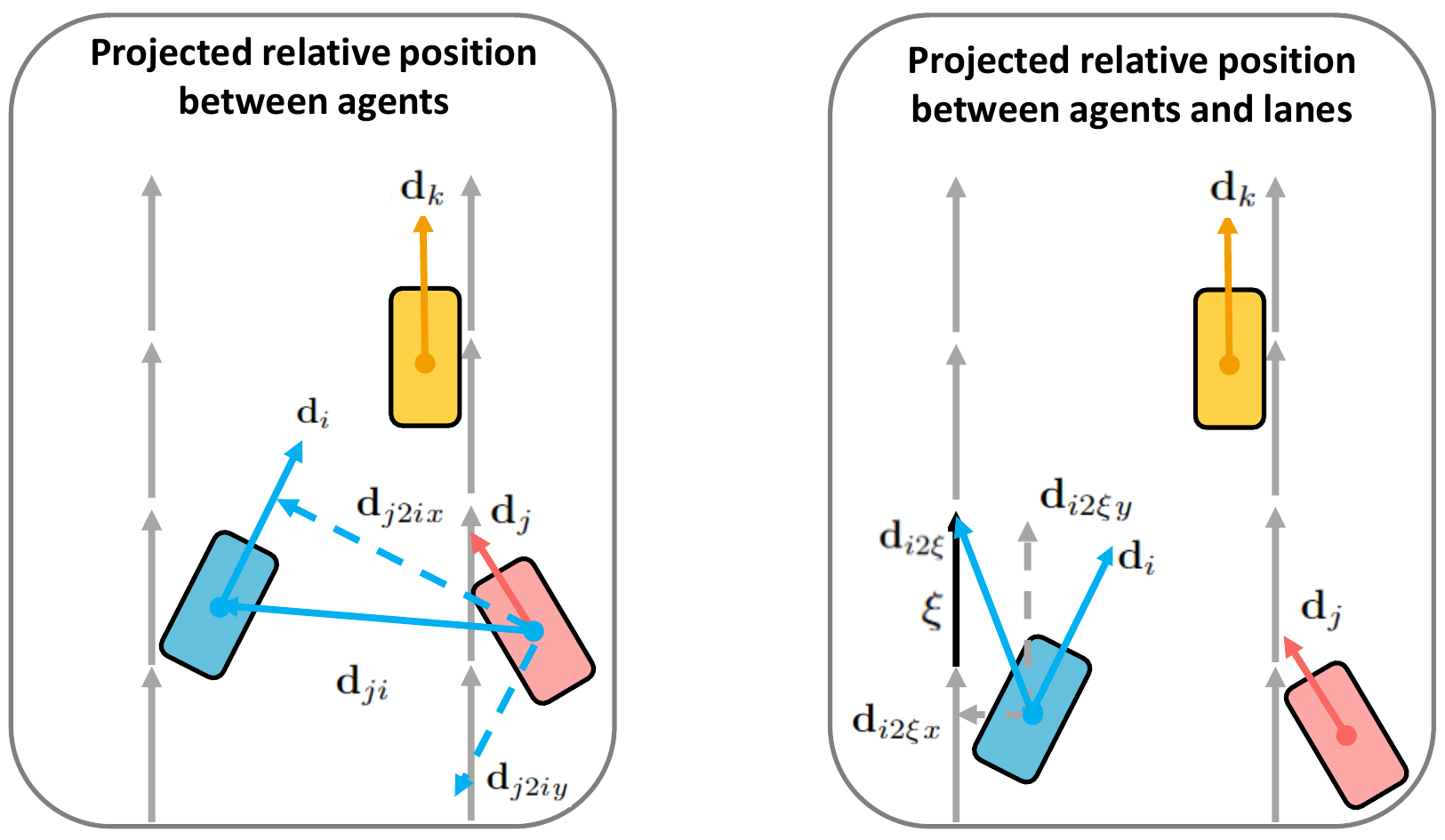} 
\caption[Projected relative vectors]{In the figure on the left, the blue agent $i$ is defined as the target agent, $\mathbf{d}_{ji}$ is the relative position vector between agent $i$ and agent $j$, $\mathbf{d}_{j2ix}$ and  $\mathbf{d}_{j2iy}$ is the lateral and longitudinal projected relative position. In the figure on the right, $\mathbf{d}_{i2\xi}$ is the relative position vector between agent $i$ and the black lane segment $\xi$, $\mathbf{d}_{i2\xi x}$ and  $\mathbf{d}_{i2\xi y}$ is the lateral and longitudinal projected relative position.} 
\label{relative_vector}
\end{figure}
\begin{align}
    &\mathbf{d}^{t}_{ij} = \mathbf{p}^{t}_j - \mathbf{p}^{t}_i &&\alpha^{t}_{ij} = \alpha^{t}_{j}-\alpha^{t}_{i}\\
    &{\color{cyan}l^{t}_{ij}} = \lVert \mathbf{d}^{t}_{ij} \rVert &&{\color{cyan}\mathbf{a}^{t}_{j2i}} = [\cos(\alpha^{t}_{ji}),\sin(\alpha^{t}_{ji})]\\
    &{\color{cyan}\mathbf{v}^{t}_{ij}} = {\color{cyan}\mathbf{v}^{t}_j} - {\color{cyan}\mathbf{v}^{t}_i} &&{\color{cyan}\mathbf{d}^{t}_{j2i}} ={\mathbf{R}^{t}_{i}}^{\top} \mathbf{d}^{t}_{ji}\\
    &{\color{cyan}s^{t}_{ij}} = \lVert {\color{cyan}\mathbf{v}^{t}_{ij}} \rVert &&{\color{cyan}\mathbf{v}^{t}_{j2i}} = {\mathbf{R}^{t}_{i}}^{\top} {\color{cyan}\mathbf{v}^{t}_{ji}}
\end{align}
where $\mathbf{d}^{t}_{ij}$ is the relative position vector between the target agent $i$ and agent $j$ at timestamp $t$, $l^{t}_{ij}$ is the distance, $\mathbf{v}^{t}_{ij}$ is the relative velocity and $s_{ij}$ is the relative speed. $\mathbf{R}^{t}_{i}$ is the rotation matrix parameterized by the heading $\alpha^{t}_{i}$ of the target agent $i$ at the timestamp $t$. $\mathbf{d}^{t}_{j2i}$ and $\mathbf{v}^{t}_{j2i}$ are the projected relative position vector and projected relative velocity respectively, and $\mathbf{a}^{t}_{j2i}$ is the relative heading vector. The projected relative position is defined as projecting relative position in the direction of the heading of the target agent. An example is shown in the left side of Fig. \ref{relative_vector}. Similarly, we obtain
the projected relative velocity by projecting relative velocity in the direction of the heading of the target agent.

We characterize the edge features between agents and lane segments as $\mathbf{e}_{al} = \{{\mathbf{R}^{T}_{i}}^{\top}\mathbf{d}^{t}_{i\xi}, {\color{cyan}\mathbf{d}^{t}_{i2\xi}},{\color{cyan}l^{t}_{i\xi}},{\color{cyan}\mathbf{v}^{t}_{i2\xi}},{\color{cyan}\mathbf{a}^{t}_{i2\xi}}|t=0,...,19; i=1,...,N_t;\xi==1,...,M\}$, where
\begin{align}
    &\mathbf{d}^{t}_{i\xi} = \mathbf{p}^0_\xi - \mathbf{p}^{t}_i &&{\color{cyan}\mathbf{v}^{t}_{i2\xi}} = {\mathbf{R}^{t}_{\xi}}^{\top} {\color{cyan}\mathbf{v}^{t}_{i}}\\
    &{\color{cyan}\mathbf{d}^{t}_{i2\xi}} = {\mathbf{R}^{t}_{\xi}}^{\top} \mathbf{d}^{t}_{i\xi} &&\alpha^{t}_{i\xi} = \alpha^{t}_{\xi}-\alpha^{t}_{i}\\
    &{\color{cyan}l^{t}_{ij}} = \lVert {\color{cyan}\mathbf{d}^{t}_{i\xi}} \rVert &&{\color{cyan}\mathbf{a}^{t}_{i2\xi}} = [\cos(\alpha^{t}_{i\xi}),\sin(\alpha^{t}_{i\xi})]
\end{align}
where $\mathbf{d}^{t}_{i\xi}$ is the relative position vector between agent $i$ and lane segment $\xi$ at timestamp $t$, $l^{t}_{i\xi}$ is the distance and $\mathbf{R}_{\xi}$ is the rotation matrix parameterized by the heading $\alpha_{\xi}$ of the target lane. $\mathbf{d}^{t}_{i2\xi}$ and $\mathbf{v}^{t}_{i2\xi}$ represent the projected relative position vector and projected relative velocity respectively, and $\mathbf{a}^{t}_{i2\xi}$ is the relative heading vector. The projected relative position between lane segment $\xi$ and agent $i$ is defined as projecting relative position in the direction of the heading of the target lane segment $\xi$. An example is shown in the right side of the Fig. \ref{relative_vector}. Similarly, we obtain the projected relative velocity $\mathbf{v}^{t}_{i2\xi}$ by projecting the agent's velocity in the direction of the heading of the target lane segment. 

The projected relative position and the projected relative velocity depict how close two individual nodes are and how fast one node moves towards another node in both lateral and longitudinal directions. Compared to relative representation, projected relative representation describes the interaction between entities in a more detailed way, allowing downstream networks to better understand their behavior. Moreover, such a projected relative representation guarantees translation invariance and rotation invariance naturally.\\

\section{PREDICTION FRAMEWORK}\label{chapter:PREDICTION FRAMEWORK}
In this section, we introduce our prediction framework TSGN which consists of a local encoder, a global encoder, and a multimodal decoder as well as the components of the local encoder. Following that, we cover details about our proposed Hierarchical Lane Transformer. 
\begin{figure}[t]
\centering
\includegraphics[width=9cm]{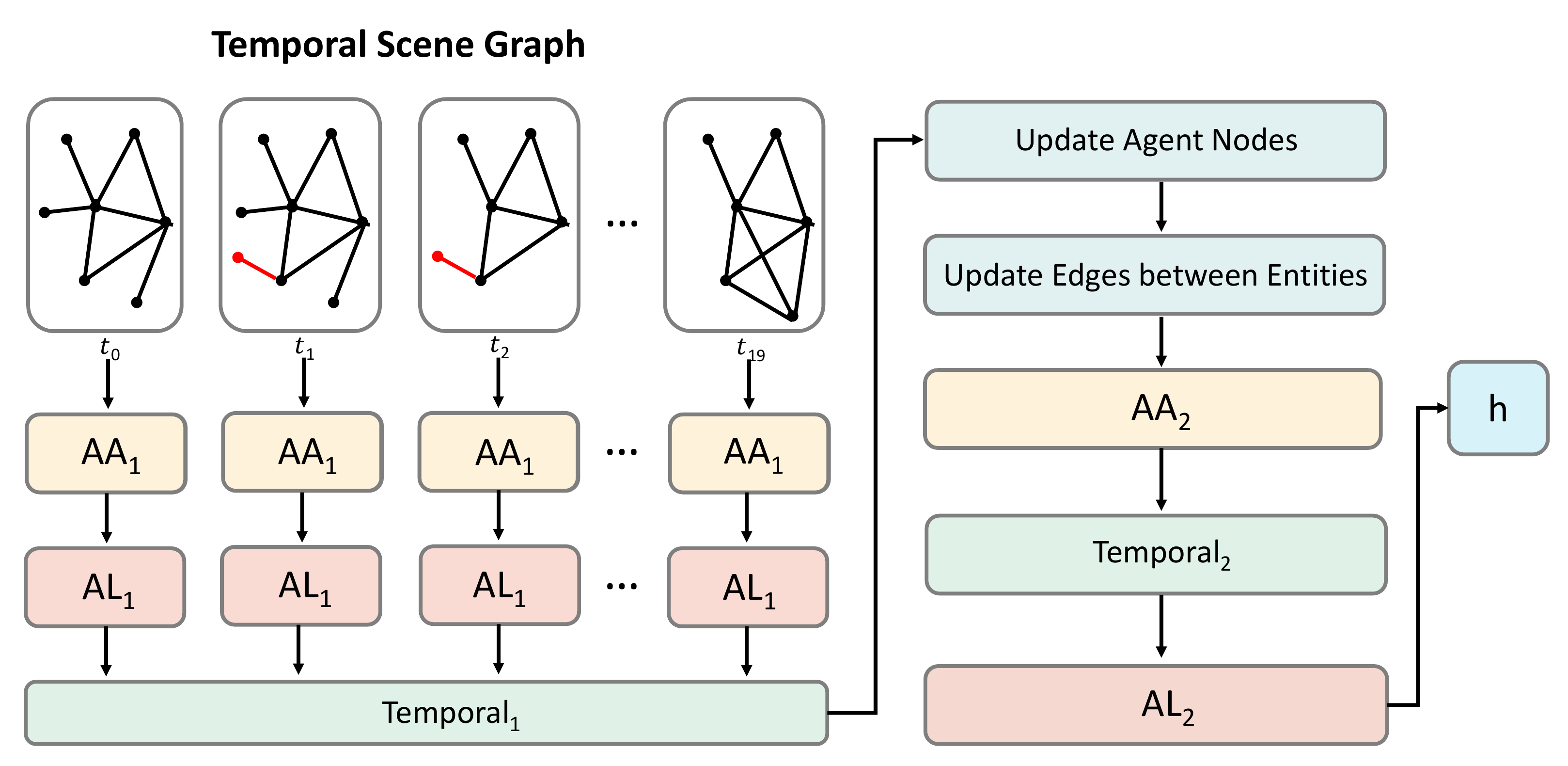}
\caption{Local Encoder of TSGN. The local encoder processes the temporal traffic scene in two stages. In the first stage, it models the agent-agent  interactions (A-A) and agent-lane interactions (A-L) per time step and uses a temporal transformer module to capture the temporal dependencies across traffic scenes. In the second stage, the refreshed agents’ features are used to update the relative spatial relationships between entities, and the agent-agent interaction is modeled again per time step. Then a temporal transformer with a classification token for summarizing the temporal information and an agent-lane transformer focusing only on the current timestamp with a larger receptive field are applied. $\mathbf{h}$ is the final local representation for all agents.}
\label{Local_Encoder}
\end{figure}
\subsection{Local Encoder} 
The local encoder operates the temporal scene graph in two stages. In the first stage, it models agent-agent interactions and agent-lane interactions per time step and uses a temporal transformer module to capture temporal dependencies across traffic scenes. Since the number of agent-lane edges is large, we use a reduced receptive field of view in this stage for efficiency. In the second stage, it uses the refreshed agents' features to update relative spatial relationships between entities and models agent-agent interactions per time step again, followed by a temporal transformer with a classification token for summarizing the temporal information and an agent-lane transformer only focusing on the current timestamp with a larger receptive field, we obtain the final updated features for each agent. In summary, the local encoder of TSGN contains the following modules: Agent-Agent Transformer in the first stage (${\rm AA}_{1}$) and second stage (${\rm AA}_{2}$), Agent-Lane Transformer in the first stage (${\rm AL}_{1}$ for all timestamps) and second stage (${\rm AL}_{2}$ only for the current timestamp), and Temporal Transformer in the first stage (${\rm Temporal}_{1}$) and second stage (${\rm Temporal}_{2}$). An overview of the local encoder is shown in Fig. \ref{Local_Encoder}. In the remainder of this subsection, we first describe the Agent-Agent Transformer, Agent-Lane Transformer, and Temporal Transformer separately, and then we detail how these transformers are going to be used in the two different stages.

\textbf{Agent-Agent Transformer} 

We use a multi-head cross-attention block to learn the interaction relationships and the importance of the influence of different surrounding agents on the center agent for each local region at each time step.
Specifically, first, we apply Multi-Layer Perceptrons (MLPs) on its node attributes and edge attributes, then we 
obtain the central agent $i$'s temporal node embedding $\mathbf{z}^t_i$ and any of its associated temporal edge embedding $\mathbf{z}^t_{ij}$ which is connected with neighboring agent $j$: 
\begin{align}
    &\mathbf{z}^{t}_{i}=\phi_{\rm{center}}\Big([{\mathbf{R}^{T}_{i}}^{\top}\mathbf{d}^{t}_i, {\color{cyan}{\mathbf{R}^{T}_{i}}^{\top}\mathbf{v}^{t}_i}, {\color{cyan}{\mathbf{R}^{T}_{i}}^{\top}\mathbf{a}^{t}_i}, {\color{cyan}s^{t}_i}, {\color{cyan}\Delta t^{t}}, \mathbf{b}_{i}] \Big)\\
    &\mathbf{z}^{t}_{ij}\!=\!\phi_{\rm{nbr}}\!\Big(\![{\mathbf{R}^{T}_{i}}^{\top}\!\mathbf{d}^{t}_{j},{\mathbf{R}^{T}_{i}}^{\top}\!\mathbf{d}^{t}_{ij},{\color{cyan}{\mathbf{R}^{T}_{i}}^{\top}\!\mathbf{v}^{t}_{ij}},\!{\color{cyan}l^{t}_{ij}},\!{\color{cyan}s^{t}_{ij}},\!{\color{cyan}\mathbf{d}^{t}_{j2i}},\!{\color{cyan}\mathbf{v}^{t}_{j2i}},\!{\color{cyan}\mathbf{a}^{t}_{j2i}},\!\mathbf{b}_{j}\!]\!\Big)
\end{align}
where $\phi_{\rm{center}}$ and $\phi_{\rm{nbr}}$ are MLP blocks. Both node and edge attributes are translation and rotation invariant, their embeddings also inherit these properties. Then, a cross attention block takes $\mathbf{z}^{t}_{i}$ as query, $\mathbf{z}^{t}_{ij}$ as key, $\mathbf{z}^{t}_{ij}$ as value, performs the scaled dot-product attention \cite{vaswani2017attention} with a gating function \cite{zhou2022hivt}
, and outputs $\hat{\mathbf{z}}^{t}_{i}$. 
We further apply an MLP block in conjunction with residual connections \cite{he2016deep} and takes $\hat{\mathbf{z}}^{t}_{i}$ as input to obtain the first updated node embedding $\mathbf{s}^{t}_{i}$ of agent $i$ at timestamp $t$ which incorporates interactions between agents.

\textbf{Agent-Lane Transformer} 

To capture implicit interactions between agents and lane segments, we employ another multi-head cross-attention block. First, we encode the local lane segments, and their associated edge attributes which are connected with the central agent $i$ by an MLP block $\phi_{\rm{lane}}$ to obtain the edge embedding: 
\begin{align}
    &\mathbf{z}_{i\xi} = \phi_{\rm{lane}}\Big([{\mathbf{R}^{T}_{i}}^{\top}\mathbf{d}_{\xi},{\mathbf{R}^{T}_{i}}^{\top}\mathbf{d}_{i\xi},{\color{cyan}\mathbf{d}_{i2\xi}}, {\color{cyan}l_{i\xi}},{\color{cyan}\mathbf{v}_{i2\xi}}, {\color{cyan}\mathbf{a}_{i2\xi}}, \mathbf{b}_{\xi}] \Big)
\end{align}
Second, we employ a multi-head cross-attention block that takes agent $i$'s node embedding $\mathbf{s}^{t}_{i}$ as query, the edge embedding $\mathbf{z}_{i\xi}$ as key and value, and performs the scaled dot-product attention the same as the Agent-Agent Transformer, followed by an MLP block in conjunction with residual connections. We then obtain the updated node embedding $\widetilde{\mathbf{s}}^{t}_{i}$ which incorporates interactions between agents and interactions between agents and lane segments. 

\textbf{Temporal Transformer}

We treat the encoded traffic scene at each timestamp as a whole and use a temporal transformer to capture temporal dependencies of all historical encoded traffic scenes. First, we add the learnable positional embeddings \cite{vaswani2017attention} to the central agent $i$'s node embedding $\{\widetilde{\mathbf{s}}^{t}_{i}\}^{T}_{t=1}$ at each timestamp and stack them into a matrix $\widetilde{\mathbf{S}}_{i}$. Then, we use a self-attention block to perform the scaled dot-product attention, in which $\widetilde{\mathbf{S}}_{i}$ is taken as query, key and value. We obtain the output of the temporal transformer $\hat{\mathbf{S}}_{i}$ which summarizes the spatial-temporal features of agent $i$ for every timestamp. 
\subsubsection{First Stage} 

The local encoder models agent-agent interactions and agent-lane interactions per time step. These two interaction modules aggregate the spatial information of the traffic scene for every timestamp. As explained before, due to the great computational complexity, we use a reduced receptive field of view in this stage for efficiency. Taking the encoded temporal traffic scene as input, the temporal transformer further summarizes the temporal information. Unlike most approaches which only capture temporal dependencies of agents' historical dynamics, we propose to incorporate agent-agent interactions and agent-lane interactions to update agents' node embedding for every timestamp, since the interactions in the past will affect the agents' pattern in the future. 

\subsubsection{Second Stage}

\label{sec:Second Part}
We use refreshed node embeddings of agents in the first stage to update the edge embeddings between agents as: 
\begin{align}
    &\mathbf{z}^{t}_{{ij}_{2}} = \phi_{\rm{nbr}_2}\Big([\hat{\mathbf{s}}^{t}_{i},\mathbf{z}^{t}_{ij}] \Big)
\end{align}
where $\{\hat{\mathbf{s}}^{t}_{i}\}^{T}_{t=1}$ is the unstacked result of $\hat{\mathbf{S}}_{i}$ and $\phi_{\rm{nbr}_2}$ is an MLP block. We use refreshed agents’ nodes embeddings $\{\hat{\mathbf{s}}^{t}_{i}\}^{T}_{t=1}$ and edges embeddings $\mathbf{z}^{t}_{{ij}_{2}}$ to repeat an agent-agent interaction encoding for every timestamp. In the first modeling of agent-agent interactions, the agents lacked information about road network, which could in turn affect their behaviors and interactions. Thus, we apply it again, after all agents are incorporated with the road network information. Afterwards, we apply another temporal transformer with an extra learnable classification token \cite{devlin2018bert} and a temporal mask that enforces the tokens only focus on the preceding time steps to summarize the whole temporal sequence into one single token $\bar{\mathbf{s}}_{i}$.
Then, the edge embeddings between agents and lane segments are calculated as: 
\begin{align}
    &\mathbf{z}_{{i\xi}_{2}} = \phi_{\rm{lane}_2}\Big([\bar{\mathbf{s}}_{i},\mathbf{z}_{i\xi}]\Big)
\end{align}
where $\phi_{\rm{lane}_2}$ is an MLP block. We input the updated extra classification token $\bar{\mathbf{s}}_{i}$ as query and updated edge embeddings $\mathbf{z}_{{i\xi}_{2}}$ between agent $i$ and lane segment $\xi$ as key and value to another same agent-lane interaction module with a larger receptive field. We obtain the final node embedding $\mathbf{h}_{i}$ of the central agent $i$. $\mathbf{h}_{i}$ summarizes the fused rich spatial-temporal features of the agent $i$. It incorporates agent $i$'s dynamics and the iterative interactions between agent $i$ and its surrounding environment. The final local representation for all agents $\mathbf{h}$ is defined as $\mathbf{h} = \{\mathbf{h}_i|i=1,...,N\}$, where $N$ is the number of agents.

\subsection{Global Encoder}
The local encoder output represents the features of the central agent within its local region, but it does not incorporate the pairwise interactions between local regions and lack the long-range dependencies in the scene. Thus, we apply a global encoder \cite{zhou2022hivt} for high-order interactions.

Similar to the Agent-Agent Transformer, we use an MLP $\phi_{rel}$ to obtain the edge embedding $\mathbf{g}_{ij}$ between agent $i$'s and agent $j$'s local regions:
\begin{align}
    &\mathbf{g}_{ij} = \phi_{\rm{rel}}\Big([{\mathbf{R}^{T}_{i}}^{\top}\mathbf{d}^{T}_{ij}, {\color{cyan}{\mathbf{R}^{T}_{i}}^{\top}\mathbf{v}^{T}_{ij}}, {\color{cyan}l^{T}_{ij}}, {\color{cyan}s^{T}_{ij}}, {\color{cyan}\mathbf{d}^{T}_{j2i}}, {\color{cyan}\mathbf{v}^{T}_{j2i}}, {\color{cyan}\mathbf{a}^{T}_{j2i}]} \Big)
\end{align}
where $T$ represents the current timestep. We take $\mathbf{h}_i$ as query, $[\mathbf{h}_j,\mathbf{g}_{ij}]$ as key and value. 
A  multi-head cross-attention block receives them as input and outputs agent $i$'s updated node embedding which incorporates global scene information. Afterwards, we employ an MLP block to map the updated node embedding of each agent to the final global representation $\widetilde{\mathbf{h}}$ with shape $[F, N, D]$, where $F$ is the number of mixture components, representing the multimodality of agents' trajectories in the future, and $D$ is the number of agents' features. 



\subsection{Multimodal Decoder}
In complex traffic scenes, agents can behave multimodally. To characterize this phenomenon, we apply a mixture density network \cite{bishop1994mixture} with GRU \cite{cho2014properties} as the decoder to output multimodal trajectories for each agent. 
Specifically, we tile the global representation $\widetilde{\mathbf{h}}$ with shape $[F, N, D]$ and the local representation $\mathbf{h}$ with shape $[N, D]$ $H$ and $F$ times, then we obtain the tiled global representation with shape $[H, F \times N, D]$ and the tiled local representation with shape $[1, F \times N, D]$, where $H$ is the number of future steps. The GRU receives the tiled global representation as inputs and the tiled local representation as the initial hidden state and outputs $\mathbf{o}$ with shape $[H, F \times N, D]$. We transpose the first dimension and second dimension of $\mathbf{o}$ and apply two  MLP blocks $\phi_{\rm{loc}}$, $\phi_{\rm{unc}}$ which take the GRU output $\mathbf{o}$ as input and output locations $\bm{\mu}$ with shape $[F, N, H, 2]$ and  associated uncertainties $\mathbf{b}$ with shape $[F, N, H, 2]$. We apply another MLP block $\phi_{\rm{prob}}$ with the concatenation of $\mathbf{h}$ and $\widetilde{\mathbf{h}}$ as input to obtain the probability $\bm{\pi}$ of possible future trajectories of all agents:
\begin{align}
    &\mathbf{o} = \rm{GRU}([\widetilde{\mathbf{h}}, \mathbf{h}]) &&\bm{\mu} = \phi_{\rm{loc}}(\mathbf{o})\\
    &\mathbf{b} = \phi_{\rm{unc}}(\mathbf{o}) &&\bm{\pi} = \phi_{\rm{prob}}([\widetilde{\mathbf{h}}, \mathbf{h}])
\end{align}
\subsection{Loss Function} 
The tracking data of Argoverse dataset $1$ contains noisy samples. Compared to Gaussian negative log-likelihood loss, Laplace negative log-likelihood loss is more robust to outliers, thus we design the multimodal decoder in that way so that it predicts a parameterization of the location and its associated uncertainty corresponding to a Laplace distribution \cite{meyer2020learning}. We use the winner-take-all strategy, and only apply the loss to the best-predicted trajectory which is defined as the one that has the minimum endpoint error. Our final loss consists of the regression loss $\mathcal{L}_{\rm{reg}}$ and the classification loss $\mathcal{L}_{\rm{cls}}$:
\begin{align}
    &\mathcal{L}_{reg} = -\frac{1}{NH}\sum^N_{i=1}\sum^{T+H}_{t=T+1}\log P\Big({\mathbf{R}^{T}_{i}}^{\top}(\mathbf{p}^{t}_{i}-\mathbf{p}^{T}_{i}) | \hat{\bm{\mu}}^{t}_{i},\hat{\mathbf{b}}^{t}_{i}\Big)\\
    &\overline{\bm{\pi}}_{i}={\rm softmax}\Big(-\sum^{T+H}_{t=T+1}\lVert {\bm{\mu}}^{t}_{i},{\mathbf{R}^{T}_{i}}^{\top}(\mathbf{p}^{t}_{i}-\mathbf{p}^{T}_{i}) \lVert \Big) \\
    &\mathcal{L}_{cls} = {\sum^N_{i=1}}{\rm CE}(\overline{\bm{\pi}}_{i}, \hat{\bm{\pi}}_{i}) \qquad \mathcal{L}=\mathcal{L}_{\rm{reg}} + \mathcal{L}_{\rm{cls}}
\end{align}
where$\hat{\bm{\mu}}^{t}_{i}$ and $\hat{\mathbf{b}}^{t}_{i}$ are the locations and their associated uncertainty of the best-predicted trajectory of agent $i$ at timestamp $t$. ${\bm{\mu}}^{t}_{i}$ is the locations of all predicted trajectories of agent $i$ at timestamp $t$. ${\rm CE}$ is the cross-entropy loss.

\subsection{Hierarchical Lane Transformer}
\begin{figure}[htbp]
\centering
\includegraphics[height=2.6cm]{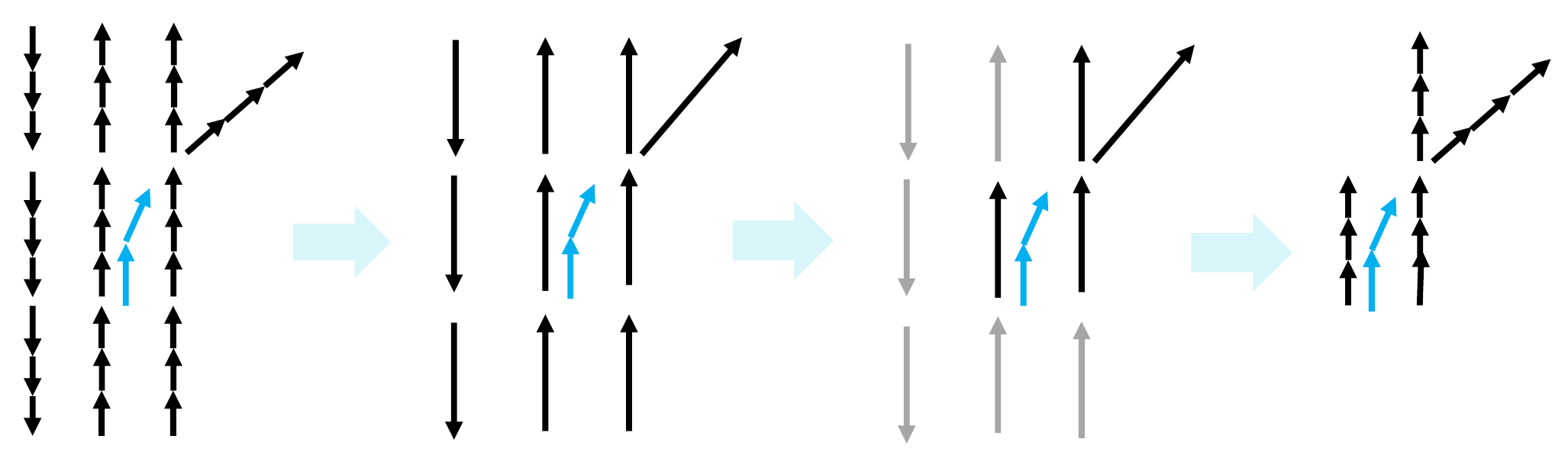}
\caption{Hierarchical process. First, the road network is constructed by lanelets instead of lane segments. Second, Agent-Lanelet Transformer selects the lanelets which score are higher than threshold. Finally, only lane segments that belong to the selected lanelets are taken into account in calculating agent-lane interactions. In this example, a lanelet consists of $3$ lane segments. The selected lanelets are shown in black, the unselected lanelets are shown in gray, and the past trajectory of the target agent is shown in blue.}
\label{hlt}
\end{figure}
The number of agent-lane edges in the local region is still very large, even if we limit the radius of the local region. When the model performs the agent-lane interaction for every timestamp, it suffers from great time complexity and space complexity. In addition, not all lane segments in the local region have an impact on the agents' future behavior. Since a lanelet \cite{bender2014lanelets} can be viewed as a set of lane segments, we can filter out the unuseful lanelets and lane segments belonging to them to reduce the computational complexity. In our case, a lanelet is defined as a sequence of $10$ centerline points.  We propose the Hierarchical Lane Transformer and apply it in the agent-lane interaction module to only select lanelets that are most important to the target agent. As the other unselected lanelets would have scarcely any influence on the target agent, this does not decrease performance. We will demonstrate it in the ablation studies.

Specifically, the Hierarchical Lane Transformer consists of an Agent-Lanelet Transformer and an Agent-Lane Transformer. The Agent-Lanelet Transformer has the same architecture as the Agent-Lane Transformer. We take the average position and the average position vector of all lane segments of lanelet $\Xi$ as the position $\mathbf{p}_{\Xi}$ and position vector $\mathbf{d}_{\Xi}$ of the lanelet $\Xi$. We create the edge embedding $\mathbf{z}_{i\Xi}$ between agent $i$ and lanelet $\Xi$ as follows:
\begin{align}
    &\mathbf{z}_{i\Xi} = \phi_{\rm{lanelet}}\Big([{\mathbf{R}^{T}_{i}}^{\top}\mathbf{d}_{\Xi},{\mathbf{R}^{T}_{i}}^{\top}\mathbf{d}_{i\Xi},\mathbf{d}_{i2\Xi},l_{i\Xi},\mathbf{v}_{i2\Xi},\mathbf{h}_{i2\Xi},\mathbf{a}_{\Xi}]\Big)
\end{align}
In the first step, the Agent-Lanelet Transformer performs the scaled dot-product attention and yields a score $\bm{\alpha}^{t}_{i\Xi}$ for each lanelet to identify the most probably lanelets that can affect the future motion of the target agent $i$.
In the second step, only lane segments whose score are greater than threshold are taken into account in calculating agent-lane interactions. To enhance the influence of the lanelet's score $\bm{\alpha}^{t}_{i\Xi}$, we multiply the scaled dot-product in Agent-Lane Transformer with $\bm{\alpha}^{t}_{i\Xi}$ before softmax function:
\begin{align}
    &\bm{\alpha}^{t}_{i\xi} = \bm{\alpha}^{t}_{i\Xi}\cdot {\rm softmax}\Big (\frac{{\mathbf{q}^{t}_i}^{\top}}{\sqrt{d_k}} \cdot \mathbf{k}^{t}_{i\xi}  \Big) 
\end{align}
where the lane segment $\xi$ belongs to the lanelet $\Xi$. An example of this hierarchical process is shown in Fig. \ref{hlt}.
\section{EXPERIMENTS}\label{chapter:EXPERIMENTS}
\subsection{Experimental Settings}
\subsubsection{Dataset}
We use the Argoverse Motion Forecasting Dataset \cite{chang2019argoverse}. It is a dataset with agent trajectories and HD map which contains $324557$ real traffic scenarios. The training, validation, and test sets contain $205942$, $39472$, and $78143$ scenarios respectively. Each scenario is a $5$-second sequence long sampled at $10$ Hz and contains the position of all agents in the past $2$ seconds. In Argoverse Motion Forecasting Challenge, we need to predict the $3$-second future positions of one target agent given the initial $2$-second observations of the scene.


\subsubsection{Metrics}
We follow the Argoverse benchmark and use minimum average displacement error (minADE), minimum final displacement error (minFDE), and miss rate (MR) to evaluate our model. These metrics allow models to forecast up to 6 trajectories for each agent. 
\subsubsection{Implementation Details}
We train all models for $64$ epochs using AdamW optimizer \cite{loshchilov2017decoupled} initialized with a learning rate of $0.0005$. We employ the cosine annealing scheduler for the learning rate decay. We set the number of layers for Agent-Agent Transformer, Agent-Lane Transformer, Agent-Lanelet Transformer, Temporal Transformer, and Global Encoder to $1$,$1$,$1$,$4$,$3$. The number of hidden units is 128, the number of heads in all multi-head attention blocks is $8$. The radius of local regions in the first stage is set to $20$ meters, while in the second stage is set to $50$ meters. We drop the agent which moved less than $6$ meters unless it is the target agent \cite{ngiam2021scene}.
\subsection{Comparison with State-of-the-art}
\begin{table}[t]
\centering
\caption{Results on Argoverse Motion Forecasting Leaderboard.}
\vspace{1em}
\begin{tabular}{c|ccc} 
\hline
Models            & minFDE & minADE & MR      \\ 
\hline
LaneGCN \cite{liang2020learning}   & 1.3640 & 0.8679 & 0.1634  \\
Scene Transformer \cite{ngiam2021scene} & 1.2321 & 0.8026 & 0.1255  \\
HOME+GOME \cite{gilles2022gohome}        & 1.2919 & 0.8904 & \textbf{0.0846}  \\
DenseTNT \cite{gu2021densetnt}          & 1.2815 & 0.8817 & 0.1258  \\
TPCN  \cite{ye2021tpcn}            & 1.2442 & 0.8153 & 0.1333  \\
MultiPath++ \cite{varadarajan2022multipath++}      & 1.214  & 0.790  & 0.13    \\
HiVT \cite{zhou2022hivt}             & 1.1693 & 0.7735 & 0.1267  \\
DCMS \cite{ye2022dcms}             & \textbf{1.1350} & 0.7659 & 0.1094  \\
Wayformer \cite{nayakanti2022wayformer}    & 1.1615 & 0.7675 & 0.1186  \\ 
\hline
TSGN              & 1.1370 & \textbf{0.7537} & 0.1223  \\
\hline
\end{tabular}
\label{Comparison with other methods}
\end{table}
We show in Tab. \ref{Comparison with other methods} the results of TSGN compared to other state-of-the-art models
on the Argoverse motion forecasting test set. The data in Tab. 1 is collected from the Argoverse leaderboard on $13/09/2022$. TSGN outperforms all the other methods in terms of minADE, and remains competitive ranking on minFDE, which verifies the superior prediction performance of our method. 
\subsection{Ablation Studies}
Our ablation studies consist of four parts: the importance of each module of TSGN, the importance of the extended scene reresentation and the importance of the Hierarchical Lane Transformer. We conduct these experiments on the Argoverse validation set.
\subsubsection{Importance of Each Module}
\begin{table}[t]
\centering
\caption{Importance of each component of our framework.}
\vspace{1em}
\setlength{\tabcolsep}{0.5mm}{
\begin{tabular}{cccccc|ccc} 
\hline
$\rm{AA}_1$ & $\rm{AL}_2$ & $\rm{Temporal}_1$ & $\rm{AL}_1$ & $\rm{AA}_2$ & $\rm{Temporal}_2$& $\rm{minFDE}$ & $\rm{minADE}$ & $\rm{MR}$  \\ 
\hline

\checkmark   &\checkmark   &\checkmark &   &  &    & $0.959$      & $0.652$     & $0.090$   \\
\checkmark  & \checkmark   & \checkmark  &\checkmark &  &  & $0.930$      & $0.644$      & $0.085$   \\
\checkmark  & \checkmark    &\checkmark  &\checkmark  & \checkmark & \checkmark& $\textbf{0.916}$      & $\textbf{0.636}$      & $\textbf{0.084}$   \\
\hline
\end{tabular}}
\label{Importance of each component of our framework}
\end{table}
 HiVT consists of ${\rm AA}_{1}$, ${\rm AL}_{2}$, ${\rm Temporal}_{1}$, and Global modules. The functions of these $4$ modules can be seen in \cite{zhou2022hivt}. As shown in Tab. \ref{Importance of each component of our framework}, ${\rm AL}_{1}$, ${\rm AA}_{2}$, and ${\rm Temporal}_{2}$ modules can improve minADE, minFDE, and MR to a certain degree. First, every moment of interaction will affect future behavior. ${\rm AL}_{1}$ captures the agent-lane interaction at each timestamp, it noticeably improves the prediction performance. Second, the components of the second stage of the local encoder (${\rm AA}_{2}$, ${\rm Temporal}_{2}$) have a significant impact on the performance. As discussed before, ${\rm AA}_{1}${} does not consider the influence that the agent-lane interaction has on the agent-agent interaction.  Based on ${\rm AA}_{1}$, ${\rm AA}_{2}$ further captures the agent-agent interaction incorporated with the map information.








\subsubsection{Importance of the Extended Scene Reresentation}
\begin{table}
\centering
\caption{Ablation studies on the extended scene representation.}
\vspace{1em}
\setlength{\tabcolsep}{1mm}{
\begin{tabular}{c|c|ccc} 
\hline
                                                      Representation method             & minFDE & minADE & MR     \\ 
\hline
\begin{tabular}[c]{@{}c@{}}raw scene representation\end{tabular}  & 0.967  & 0.662  & 0.092  \\ 
\hline
\begin{tabular}[c]{@{}c@{}}extended scene representation\end{tabular}  & \textbf{0.959}  & \textbf{0.652}  & \textbf{0.090}  \\
\hline
\end{tabular}}
\label{Importance of the extended scene reresentation}
\end{table}
In this ablation study, we use HiVT \cite{zhou2022hivt} as the prediction framework. As shown in Tab. \ref{Importance of the extended scene reresentation}, compared to the original representation, the extended scene representation yields better performance. Our proposed representation provides more information about the entities and more finely describes the relative relationships between these entities, allowing downstream networks to better understand the interactions between individuals in a complex scene. 
\subsubsection{Importance of the Hierarchical Lane Transformer}
\begin{table} 
\centering
\caption{Ablation studies on the lanelets selection module.}
\vspace{1em}
\setlength{\tabcolsep}{0.6mm}
\begin{tabular}{c|ccc|c|c} 
\hline
                                                    A-L module             & minFDE & minADE & MR     & Lane usage rate  &Speed (ms)\\ 
\hline
\begin{tabular}[c]{@{}c@{}}w/o Lanelet selection\end{tabular}  & \textbf{0.916}  & \textbf{0.636}  & \textbf{0.084} & 100\% &331              \\ 
\hline
\begin{tabular}[c]{@{}c@{}}w/ Lanelet selection\end{tabular}  & 0.922  & 0.638  & 0.085 & \textbf{45.8\%} &\textbf{226}           \\
\hline
\end{tabular}
\label{Importance of the Hierarchical Lane Transformer}
\end{table}
In this ablation study, we use TSGN as the prediction framework. We present the performance evaluation of the Hierarchical Lane Transformer in Tab. \ref{Importance of the Hierarchical Lane Transformer}. In practice, we select the lanelets whose score $\bm{\alpha}^{t}_{i\Xi}$ is greater than $0.75 \cdot \overline{\bm{\alpha}^{t}_{i\Xi}}$, where $\overline{\bm{\alpha}^{t}_{i\Xi}}$ represents the scatter mean score of all lanelets which have a shared edge with agent $i$. 
We measure the resulting average proportion of selected lanes to be $45.8\%$, dividing the number of processed lane features by more than 2. As a consequence, the computational complexity of agent-lane interaction is significantly reduced and the measured inference speed improves greatly. Moreover, the Hierarchical Lane Transformer also keeps the prediction performance. As shown in Tab. \ref{Importance of the Hierarchical Lane Transformer}, the minFDE, minADE and MR are barely significantly worse. 

\subsection{Qualitative Results of TSGN}
%
\begin{figure}[t]
\centering
\subfigure[]{
\centering
  \includegraphics[height=3.5cm,width=3.5cm]{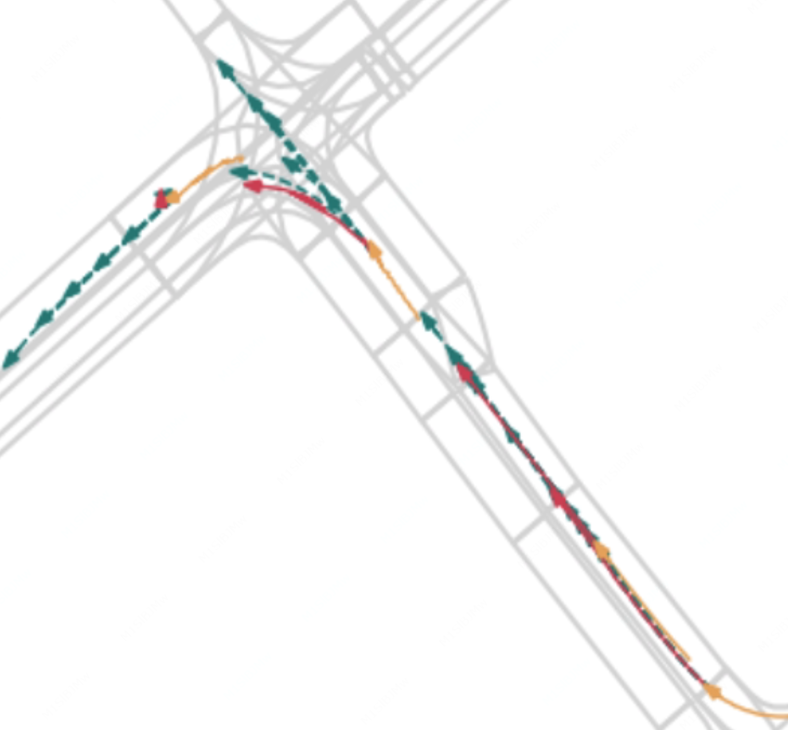} 
  \label{fig:Qualitative Results of TSGN 1}}
\subfigure[]{
\centering
  \includegraphics[height=3.5cm,width=3.5cm]{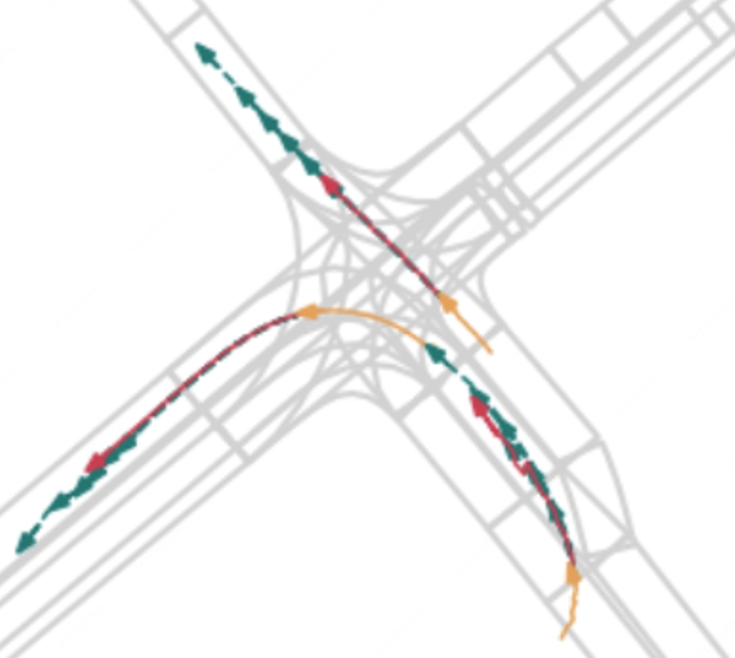}
  \label{fig:Qualitative Results of TSGN 2}}

\subfigure[]{
\centering
    \includegraphics[height=3.5cm,width=3.5cm]{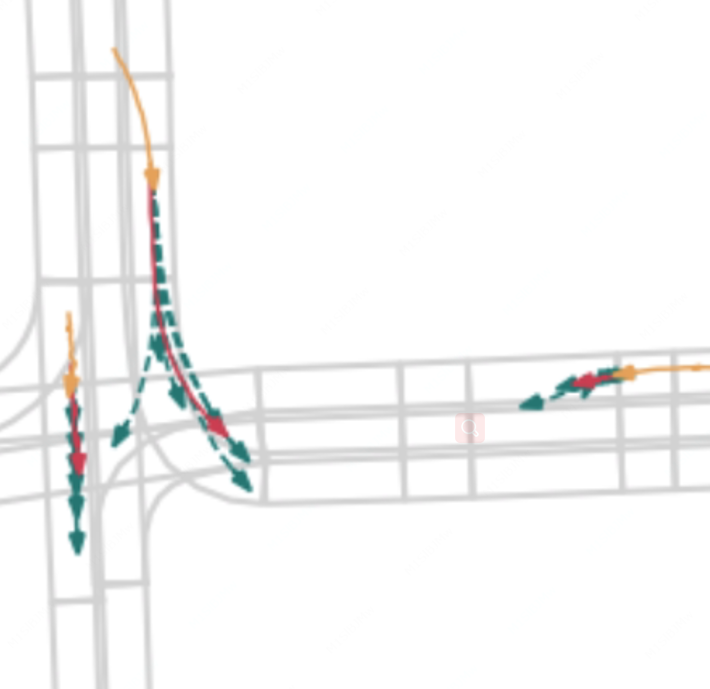} 
  \label{fig:Qualitative Results of TSGN 4}}
\subfigure[]{
\centering
  \includegraphics[height=3.5cm,width=3.5cm]{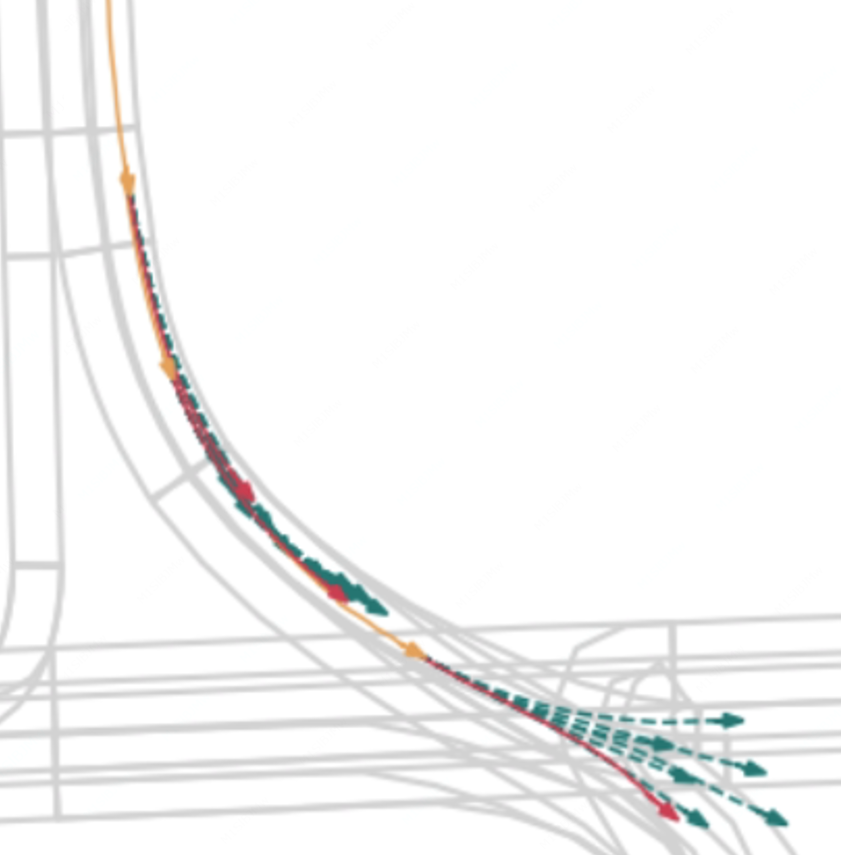}
  \label{fig:Qualitative Results of TSGN 5}}
\caption{Qualitative Results of TSGN. For clarity, we visualize only agents which are observed at current timestamp and move more than $6$ meters. We present the past trajectories in orange, the ground-truth trajectories in red and the predicted trajectories in green.}
\label{Qualitative Results of TSGN}
\end{figure}

 From Fig. \ref{fig:Qualitative Results of TSGN 1} to Fig. \ref{fig:Qualitative Results of TSGN 5}, we visualize the qualitative results of TSGN in $4$ traffic scenes. One can conclude that TSGN can make multimodal predictions for all agents simultaneously and the prediction in each maneuver is reasonable. Comparing the best-predicted trajectory with the ground truth, we see that TSGN can make accurate predictions. 

\subsection{Comparison with HiVT in Worse Cases}
\label{Comparison with HiVT in Worse Cases}
\begin{figure}[t]
\centering
\subfigure[]{
\centering
  \includegraphics[height=3.5cm,width=3.5cm]{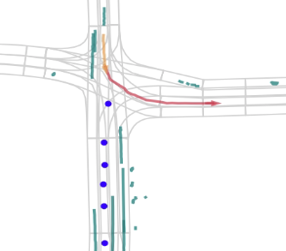} 
  \label{Comparison 1b}}
\subfigure[]{
\centering
  \includegraphics[height=3.5cm,width=3.5cm]{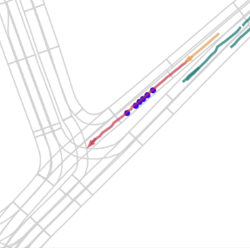}
  \label{Comparison 2b}}

\subfigure[]{
\centering
    \includegraphics[height=3.5cm,width=3.5cm]{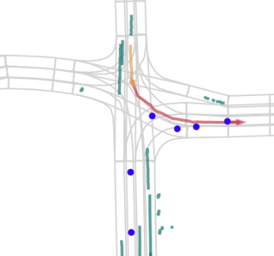} 
  \label{Comparison 1}}
\subfigure[]{
\centering
  \includegraphics[height=3.5cm,width=3.5cm]{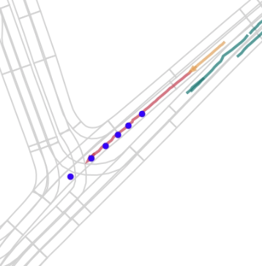}
  \label{Comparison 2}}
\caption[Comparison results]{Qualitative results of HiVT and TSGN are shown in the first row and second row respectively. The central agent's past trajectory is shown in orange, the ground-truth trajectory is shown in red, and the endpoints of predicted trajectories are shown in blue. The past trajectories of other agents are shown in green.}
\label{Comparison results}
\end{figure}
To compare with HiVT, we present the qualitative results of TSGN and HiVT in some worse cases of HiVT, in which the minFDE of HiVT is greater than $10$ meters. For clarity, we only visualize end points of the $6$ predicted trajectories of the central agent. 
We compare these two models in two different traffic scenes. Fig. \ref{Comparison 1b} and Fig. \ref{Comparison 2b} show the results of HiVT, while Fig. \ref{Comparison 1} and Fig. \ref{Comparison 2} show the results of TSGN. From Fig. \ref{Comparison 1b} and Fig. \ref{Comparison 1}, we observe that TSGN can more successfully predict reasonable multimodal future trajectories for agents. In addition, as shown in Fig. \ref{Comparison 2b} and Fig. \ref{Comparison 2}, TSGN is more sensitive to speed-related information. When the agent is accelerating or decelerating, it can better perceive the current motion state of the agents. 

\subsection{Failed Cases}
\begin{figure}[t]
\centering
\subfigure[]{
\centering
  \includegraphics[height=3.5cm,width=3.5cm]{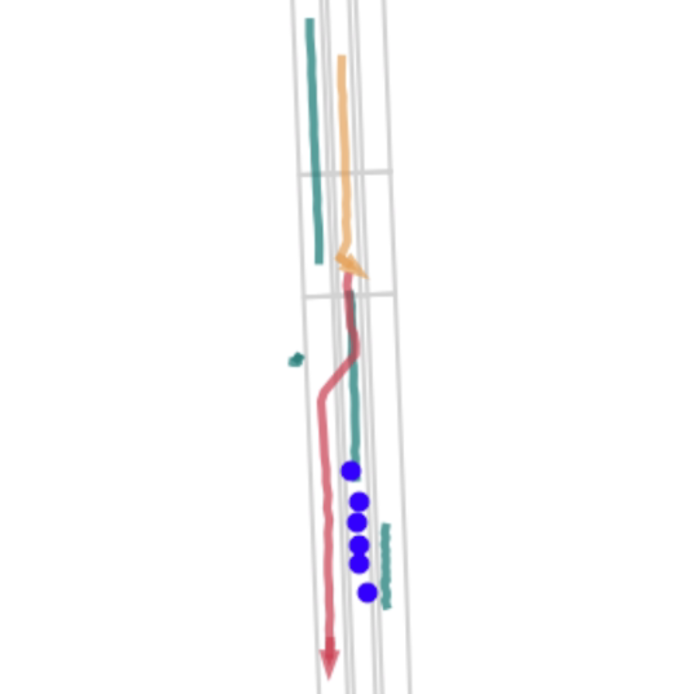}
  \label{failed case 1}}
\subfigure[]{
\centering
  \includegraphics[height=3.5cm,width=3.5cm]{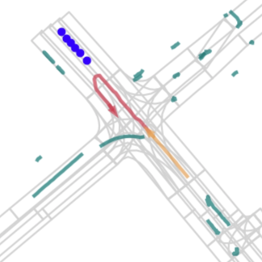}
  \label{failed case 3}}
\subfigure[]{
\centering
    \includegraphics[height=3.5cm,width=3.5cm]{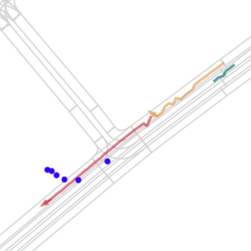} 
  \label{failed case 4}}
\subfigure[]{
\centering
  \includegraphics[height=3.5cm,width=3.5cm]{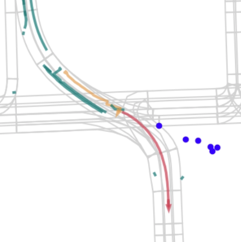}
  \label{failed case 6}}
\caption[]{Failed cases of TSGN. The visualization method is the same as Fig.\ref{Comparison results}.}
\label{Failed cases of TSGN}
\end{figure}
From Fig. \ref{failed case 1} to Fig. \ref{failed case 6}, we present some classic failed cases of TSGN. First, as shown in Fig. \ref{failed case 1} and Fig. \ref{failed case 3}, we conclude that it’s hard for TSGN to make accurate predictions when the agents intend to perform maneuver 'lane change' or 'U turn'. One possible cause of the failure is, compared with conventional maneuvers, such as 'Go straight', 'left turn' and 'right turn', the proportion of maneuver 'lane change' and 'U-turn' is very low. It is difficult for the model to learn to predict these kinds of intentions. 
Second, in Fig. \ref{failed case 4}, we observe that the prediction performance of TSGN is not robust, if the tracking information of the agent in the past is unstable.  The prediction results seem to be overfitting to past perception errors. 
Third, when the agent moves in a complex traffic scene where the surrounding lane segments are extraordinarily tangled, e.g., the traffic scene in Fig. \ref{failed case 6}, TSGN fails in understanding the map structure and makes unreasonable multimodal predictions. We want TSGN to output distinct maneuvers, but it only presents predictions in one maneuver.
\subsection{Visualization of Selected Lanelets by Hierarchical Lane Transformer and Their Scores}
\begin{figure}[t]
\centering
\subfigure[]{
\centering
  \includegraphics[height=4cm,width=4cm]{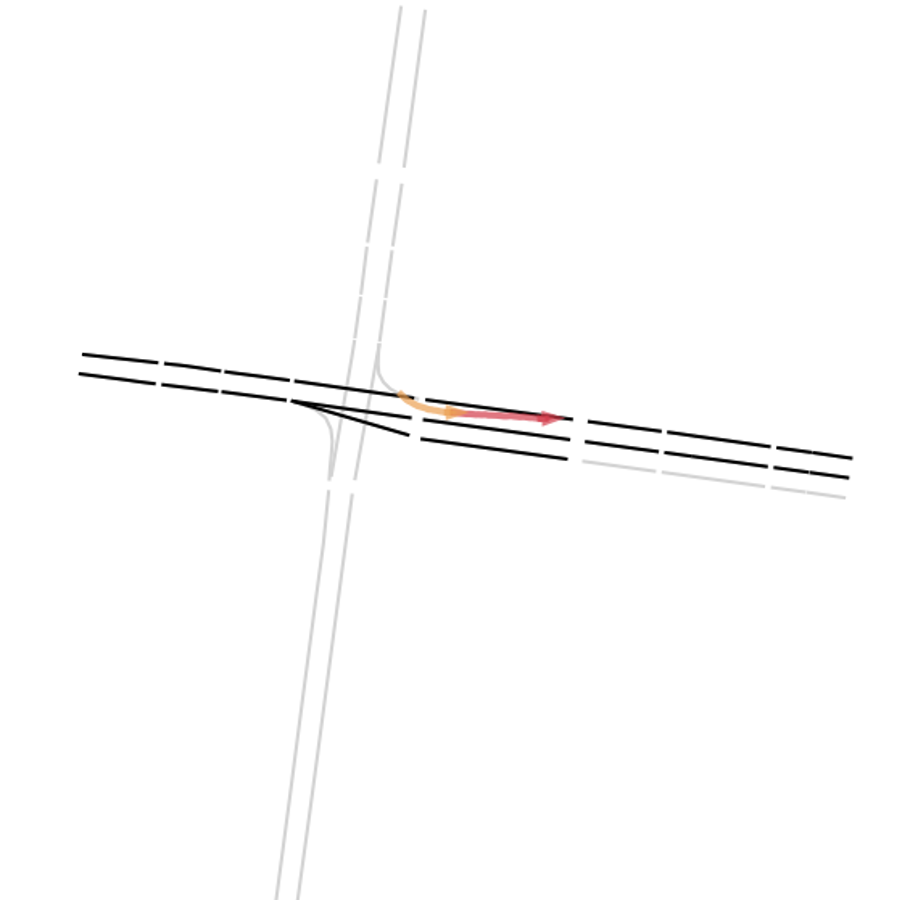} 
  \label{selected lanelet 1}}
\subfigure[]{
\centering
  \includegraphics[height=4cm,width=4cm]{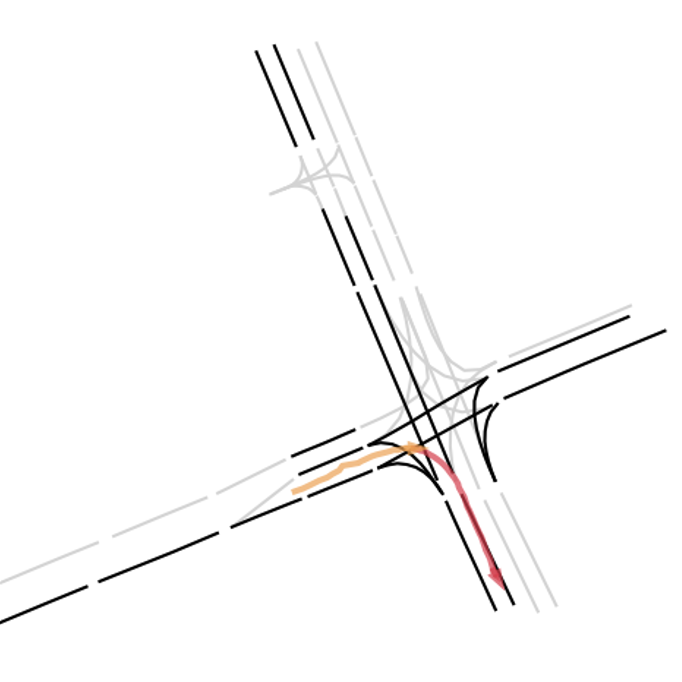}
  \label{selected lanelet 2}}
\subfigure[]{
\centering
    \includegraphics[height=4cm,width=4cm]{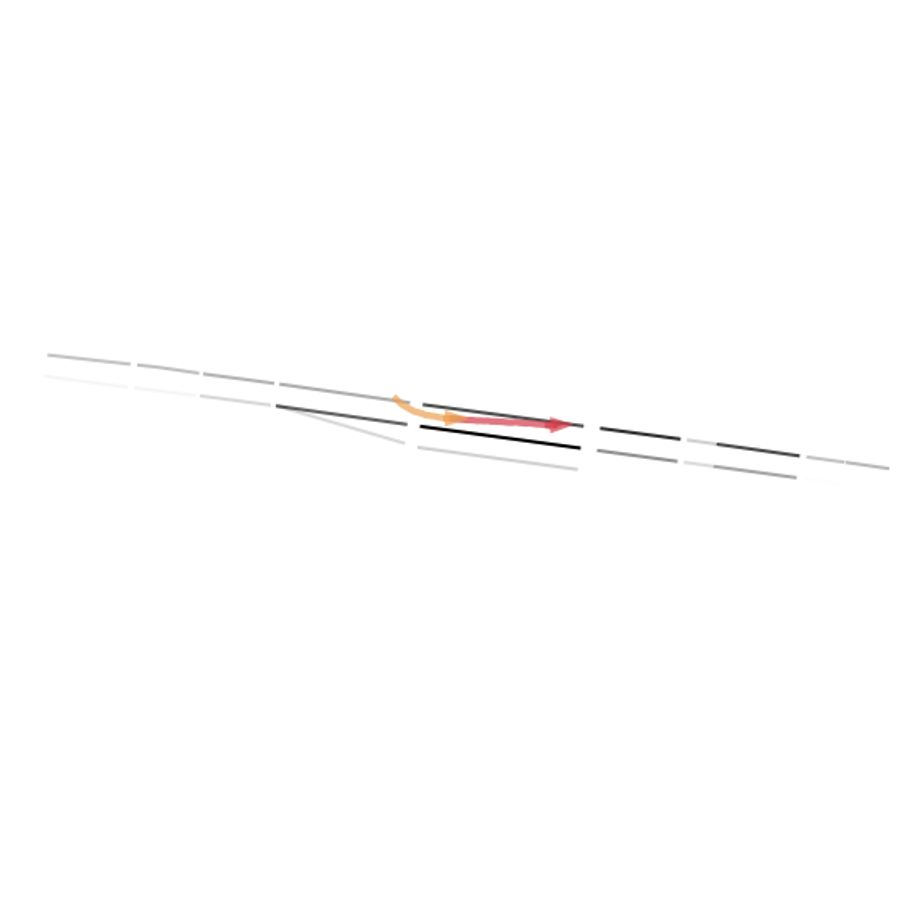} 
  \label{lanelet's score 1}}
\subfigure[]{
\centering
  \includegraphics[height=4cm,width=4cm]{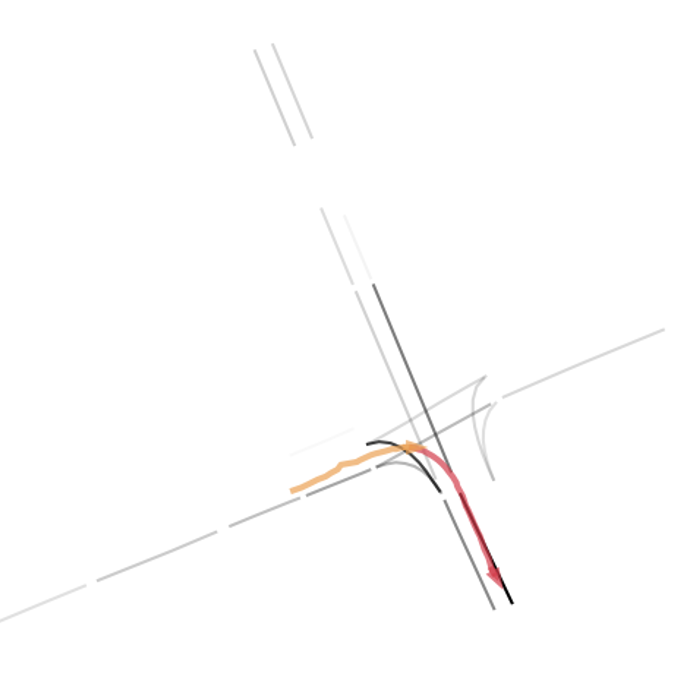}
  \label{lanelet's score 2}}
\caption{Selected lanelets by Hierarchical Lane Transformer and their’ scores. The agent's past trajectory is shown in orange, and the ground-truth trajectory is shown in red. In the first row, selected lanelets and unselected lanelets are presented in black and grey respectively. In the second row, the depth of the selected lanelet’s color represents the level of the score. The higher the score, the more important the lanelet is to the target agent.}
\label{Selected lanelets}
\end{figure}

In Fig. \ref{selected lanelet 1} and \ref{selected lanelet 2}, we present the selected lanelets by Hierarchical Lane Transformer of two different traffic scenes. Fig. \ref{lanelet's score 1} and \ref{lanelet's score 2} show the corresponding selected lanelets' scores. We observe that our model can select the most important lanelets to the target agent. The selected lanelets contain the possible future trajectories of the target agent. In addition, the prediction of lanelets’ scores is reasonable. 

  

\section{CONCLUSION}\label{chapter:CONCLUSION}

In this paper, we propose TSGN with projected vectorized representation for multi-agent trajectory prediction. It models the temporal traffic scene as a temporal scene graph and uses diverse transformer-based interaction modules and temporal dependencies modules to operate this temporal scene graph. Moreover, we present a Hierarchical Lane Transformer module inside TSGN, which effectively and efficiently captures the influence of the HD-map on the future motion of target agents. Experiments show our method achieves state-of-the-art performance on the Argoverse motion forecasting benchmark. 
\bibliographystyle{IEEEtran}
\bibliography{IEEEabrv,mybib}
\end{document}